# Uncertainty Quantification in Multivariable Regression for Material Property Prediction with Bayesian Neural Networks

*Longze Li[1], Jiang Chang[1], Aleksandar Vakanski[1]\*, Yachun Wang[2], Tiankai Yao[3], Min Xian[1]*

[1] Department of Computer Science, University of Idaho, Idaho Falls, ID 83404, USA
[2] Nuclear Science & Technology (NS&T), Idaho National Laboratory, Idaho Falls, ID 83415, USA
[3] Materials & Fuels Complex (MFC), Idaho National Laboratory, Idaho Falls, ID 83415, USA

**Abstract**

With the increased use of data-driven approaches and machine learning-based methods in material science, the importance of reliable uncertainty quantification (UQ) of the predicted variables for informed decision-making cannot be overstated. Accordingly, various approaches for UQ have been developed, either through a quantified measure of the variance in the target variable, vie confidence intervals, or by other means. Among the conventional UQ methods for multivariable regression, Gaussian Process Regression (GPR) has been generally adopted as the state-of-the-art approach that provides accurate single-point predictions and reliable uncertainty estimates. However, GPR has also important limitations, since the commonly used isotropic covariance kernels, such as Gaussian and Matern kernels, are less suitable for modeling functions with spurious covariates or anisotropic smoothness (e.g., in material property prediction, small changes in the microstructure can result in significant variations in material properties). In addition, UQ in material property prediction poses unique challenges, including difficulties in quantifying uncertainties due to the multi-scale and multi-physics nature of advanced materials, intricate interactions between numerous factors, limited availability of large curated datasets for model training, substantial variability in material properties due to test conditions and inherent factors, measurement errors, and others. Recently, Bayesian Neural Networks (BNNs) have emerged as a promising approach for UQ, offering a probabilistic framework for capturing uncertainties within neural networks. In this work, we introduce an approach for UQ within physics-informed BNNs, which integrates knowledge from governing laws in material modeling to guide the models toward physically consistent predictions. To evaluate the effectiveness of this approach, we present case studies for predicting the creep rupture life of steel alloys. Experimental validation with three datasets of collected measurements from creep tests demonstrates the ability of BNNs to produce accurate point and uncertainty estimates that are competitive or exceed the performance of GPR. Similarly, we evaluated the suitability of BNNs for UQ in an active learning application and reported competitive performance. The most promising framework for creep life prediction is BNNs based on Markov Chain Monte Carlo approximation of the posterior distribution of network parameters, as it provided more reliable results in comparison to BNNs based on variational inference approximation or related NNs with probabilistic outputs. The codes are available at: https://github.com/avakanski/Creep-uncertainty-quantification.

## 1. Introduction

*Uncertainty Quantification* (UQ) plays a crucial role in various science and engineering disciplines. In the field of material science, the application of computational modeling has significantly accelerated the discovery of novel materials with enhanced properties. Determining the level of confidence in the predictions made by computational models is of high importance, as high levels of uncertainty can result in large deviations from the actual material behavior in practical applications [1]. With the increasing complexity of computational modeling, the computational cost associated with numerical UQ models has also increased, necessitating the development of computationally efficient methods for both prediction and uncertainty estimates [2]. In general, uncertainties can be categorized into two main types: aleatoric uncertainty, which arises due to the inherent process randomness (e.g., similarities in experimental data from the same experiment), and epistemic uncertainty, related to the discrepancies due to lack of training data or imperfections in the computational models [3–4].

Recent advancements in Artificial Intelligence, particularly in Machine Learning (ML) and Artificial Neural Networks, ushered in a new era for design of experiments and materials modeling [5]. Among the *conventional ML models* with an inherent ability for UQ in regression tasks are Quantile Regression (QR) [6], Gaussian Process Regression (GPR) [7], and Natural Gradient Boosting (NGBoost) [8]. QR and NGBoost have shortcomings due to the lack of closed-form parameters estimation and are prone to overestimating the uncertainty level in data [9]. In most related works, GPR is generally reported as the state-of-the-art approach for UQ in multivariable regression and it stands out for its predictive accuracy and uncertainty estimates. On the other hand, GPR also has important limitations, since the commonly employed kernels (such as Gaussian and Matern kernels) based on isotropic covariance functions have continuous sample paths, and are therefore less suitable for material properties prediction, where small changes in the microstructure can result in significant changes in material properties [10]. In addition, isotropic covariance kernels are less suitable for modeling functions with spurious covariates or anisotropic smoothness [10]. Although researchers have proposed advanced GPR alternatives (e.g., sparse additive GPR [11]), they typically introduce novel challenges and require finetuning of additional hyperparameters.

In recent years, *artificial Neural Networks* (NNs) have demonstrated remarkable success in both classification and regression tasks dealing with high-dimensional non-linear data [12]. Whereas NNs for classification tasks inherently output the probabilities in the model's prediction for each class, traditional NNs for regression tasks typically output only *single-point predictions* of the target variables (commonly referred to as *point estimates*). To provide uncertainty assessments, previous works proposed approaches such as Deep Ensembles [13] or Monte Carlo (MC) Dropout [14]. These methods introduced modifications in the outputs of traditional NNs with deterministic parameters to generate probabilistic predictions, allowing for UQ. Importantly, NNs with stochastic parameters, referred to as *Bayesian NNs* (BNNs) [15–17], have emerged as a promising approach for UQ that provides a probabilistic framework for capturing uncertainties in data-driven NN models. In comparison to conventional ML approaches for UQ, BNNs offer substantial flexibility in terms of the model structure, size, and parameter settings, and have the potential for efficient and reliable UQ modeling [18].

In this work, we propose an approach for predicting creep rupture life in steel alloys using physics-informed BNNs. The approach incorporates physics-informed features based on governing creep laws into BNNs to estimate the uncertainties in the model's prediction of rupture life. The effectiveness of the proposed approach is additionally evaluated in the context of active learning (AL) [19]. By combining the variance reduction technique with *k*-mean clustering for selecting the most uncertain and diverse data points for training a model, we introduce a trade-off between exploration and exploitation of the solution space in AL. We conducted experimental validation with three datasets, consisting of collected data from creep tests with Stainless-Steel 316 alloys, Nickel-based superalloys, and Titanium alloys. The considered implementations of BNNs include networks employing Variational Inference (VI) and Markov Chain Monte Carlo (MCMC) approximation of the posterior distribution of the network parameters. We also evaluated the performance of traditional UQ regression models QR, NGBoost, and GPR, as well as deterministic NNs with point estimates and probabilistic outputs (Deep Ensembles, MC Dropout). The experimental results with a set of predictive single-point and uncertainty metrics demonstrate that MCMC BNNs are the most promising UQ method for creep rupture life prediction, with performance that is competitive or exceeds the performance of GPR. The results also demonstrate that physics-informed knowledge leverages the models' capacity for improved creel life prediction.

Although prior works have explored the application of ML approaches for predicting material properties [20–26] and uncertainty estimates in the predictions [16][27–29], our proposed approach introduces novel concepts related to BNNs with incorporated physics priors for UQ in material property prediction. Specifically, our work was inspired and has similarities to the recent article by Mamun et al. [9], who proposed an approach for predicting the creep rupture life of ferritic steels using conventional ML methods. The authors employed GPR to calculate point estimates and uncertainty estimates in the predicted rupture lire. Differently from the work by Mamun et al. [9], we developed approaches for point regression and UQ in predicting creep rupture life based on BNNs, and we demonstrated that Bayesian deep learning models consistently achieved improved performance in comparison to GPR for this task. Similarly, a body of work in the literature utilized Physics-Informed ML (PIML) to integrate knowledge from governing physics laws and data-driven methods for obtaining more consistent predictions [30–34]. In our proposed approach, we drew inspiration from the work by Zhang et al. [31], where the authors introduced physics-informed features and a physics-informed NN loss for predicting creep rupture life. On the other hand, the authors did not consider UQ in their work, as well as, their focus was on designing standard NNs with deterministic parameters for creep life prediction. In another related paper, Oliver et al. [16] investigated the use of BNNs for UQ in the field of material science, and they developed VI-based ensemble methods for predicting the properties of composite materials. Differently from our work, the authors in [16] did not consider the integration of physics-informed knowledge in BNNs, and also they used simulated data as a proof-of-concept for the proposed approaches, whereas we used collected data from creep tests for experimental validation. Similarly, researchers have proposed incorporating physics-informed priors into BNNs in prior works [35][36], however, these works focus on other tasks, and to the best of our knowledge, this is the first work to apply such framework for material property prediction. Lastly, although many previous works have studied AL to prioritize the most informative sample for model training [9][35–38], in this paper we showed that physics-informed BNNs have the potential to accelerate the model training in AL for material property prediction.

The contributions of this paper include:
- Introduced a physics-informed BNNs approach for material property prediction, that incorporates physics knowledge for guiding the solutions of BNNs.
- Performed a comprehensive evaluation of the performance of conventional ML models, traditional NNs with probabilistic outputs, and BNNs for creep rupture life prediction and uncertainty estimation.
- Applied the UQ frameworks for an active learning task to iteratively select data points with the highest epistemic uncertainty and diversity for faster model training with fewer data points.

The paper is organized as follows. Section 2 provides a problem definition and presents an overview of ML methods for UQ in multivariable regression. Section 3 introduces the proposed approach of physics-informed BNNs and describes the active learning application. Experimental validation and comparative analysis are provided in Section 4. Section 5 provides a discussion of the work, and the last section concludes the paper.

## 2. Preliminaries: Frameworks for Uncertainty Quantification

The considered problem is a multivariable regression task, where based on a set of $N$ observed data points $\mathbf{X} = \{(\mathbf{x}_1, \mathbf{x}_2, \ldots, \mathbf{x}_N), \mathbf{x}_i \in \mathbb{R}^d\}$, the goal is to estimate the values of a target variable $\mathbf{Y} = \{(y_1, y_2, \ldots, y_N) \in \mathbb{R}\}$. The target variable $\mathbf{Y}$ is a material property that we are interested in. In this paper, we consider creep rupture life as a property of interest. The observed data points $\mathbf{X}$ provide relevant material information, such as composition, known physical, mechanical, or other properties, and experimental conditions (such as temperature, and stress level) that are important for estimating the target variable.

For a training dataset $\mathcal{D} = \{(\mathbf{x}_i, y_i)\}_{i=1}^{N}$ and a new data point $\mathbf{x}^*$ which does not belong to the previously observed data $\mathbf{X}$, the objective is to find a mapping function $f$ that estimates the target value, i.e., $y^* = f(\mathbf{x}^*)$. The value $y^*$ is referred to as *single-point prediction* or *point estimate*. In addition, the focus of this paper is on methods that provide *uncertainty quantification* for the predicted value $y^*$, either through a quantified measure of the variance of $y^*$, via confidence intervals, or by other means. The next sections provide an overview of conventional ML and DL frameworks for UQ in regression tasks.

### 2.1 Conventional Machine Learning Methods

Whereas many ML models for classification tasks inherently output the probabilities in the model's prediction for each class, most ML models for regression tasks typically provide only point estimates of the target value. Consequently, several approaches have been developed that employ or modify standard regression models in order to provide uncertainty estimates.

*2.1.1 Quantile Regression*

Quantile Regression (QR) [6] is a non-parametric approach for estimating the conditional quantiles—and therefore, uncertainties—in prediction variables. For a quantile $\tau$, the conditional quantile function of a target variable $\mathbf{Y}$ given observed data $\mathbf{X}$ in QR is defined as:

$$Q_{\mathbf{Y}|\mathbf{X}}(\tau) = \inf\{y \in \mathbb{R} | \mathcal{P}(\mathbf{Y} \leq y | \mathbf{X} = \mathbf{x}) \geq \tau\} \text{ for } \tau \in (0,1) \ . \tag{1}$$

The QR loss function for quantile $\tau$ is as follows:

$$\mathcal{L}_\tau = \sum_{i: y_i \geq Q_{y_i|\mathbf{x}_i}(\tau)}^{N} \tau |y_i - Q_{y_i|\mathbf{x}_i}(\tau)| + \sum_{i: y_i < Q_{y_i|\mathbf{x}_i}(\tau)}^{N} (1-\tau) |y_i - Q_{y_i|\mathbf{x}_i}(\tau)| \ . \tag{2}$$

For different quantiles, QR fits a separate model to minimize the loss function in (2). For instance, for a new data point $\mathbf{x}^*$, by estimating the quantile functions $Q_{y^*|\mathbf{x}^*}(\tau = 0.025)$ and $Q_{y^*|\mathbf{x}^*}(\tau = 0.975)$, we can obtain the 95% prediction interval of the target variable around the point estimate $y^*$. Similarly, setting the quantile to $\tau = 0.5$, QR solves median regression and outputs the median values.

Compared to traditional regression methods that only estimate the conditional mean of $y^*$ given $\mathbf{x}^*$, QR estimates the conditional distribution $Q_{y^*|\mathbf{x}^*}(\tau)$, inherently providing uncertainty estimation. Other advantages of QR include: robustness to outliers, there are no distributional assumptions resulting in distribution-free estimation and inference, and it can be used with any base model by replacing the original loss function with the quantile loss function.

*2.1.2 Natural Gradient Boosting Regression*

Natural Gradient Boosting (NGBoost) Regression [8] is a probabilistic variant of the traditional Gradient Boosting method [39]. Gradient Boosting is an ensemble learning technique that combines a collection of sequentially-trained base learners (such as Decision Trees, or other ML models) into a strong learner. In each iteration of the training process, the base learners are trained to fit the residuals of the errors from the previous iteration, and the ensemble iteratively minimizes the prediction error.

NGBoost method comprises three key components: a collection of base learners, parametric form of the conditional distribution $\mathcal{P}(\mathbf{Y}|\mathbf{X},\theta)$, and scoring mechanism. The scoring mechanism ensures that the predicted distribution closely aligns with the actual distribution. To quantify the agreement, the negative likelihood is typically employed for scoring computation. For a new data point $\mathbf{x}^*$, the point estimate of the target variable $y^*$ and the uncertainty quantified as the standard deviation $\sigma^*$ are obtained from the conditional distribution $\mathcal{P}(y^*|\mathbf{x}^*,\theta)$. Advantages of NGBoost include the inherent property of uncertainty quantification, as well as the flexibility to be used with any base learners, and any distributions with continuous parameters and scoring rules.

*2.1.3 Gaussian Process Regression*

Gaussian Process Regression (GPR) [7][40] is a non-parametric Bayesian approach suitable for performing both function approximation and uncertainty estimation. Accordingly, GPR represents a collection of random variables as a multivariate Gaussian distribution over a set of data points.

For a Gaussian Process $\mathcal{N}(\mathbf{Y}|\boldsymbol{\mu},\mathbf{K})$, $\mathbf{Y}$ denotes a vector of function values estimated at $n$ data points $[f(\mathbf{x}_1), \ldots, f(\mathbf{x}_n)]$, $\boldsymbol{\mu}$ is the mean of the Gaussian Process that by default is assigned to be zero, and $\mathbf{K}$ is a positive definite covariance matrix. The smoothness of the distribution across functions is determined by the covariance kernel $K_{i,j} = k(\mathbf{x}_i, \mathbf{x}_j)$ that defines the covariance between the function values $f(\mathbf{x}_i)$ and $f(\mathbf{x}_j)$.

Several kernel functions are used in practice, parameterized by a set of hyperparameters. For instance, the Radial Basis Function (RBF) is among the most common kernel functions, and it is defined as:

$$k(\mathbf{x}_i, \mathbf{x}_j) = \sigma_f^2 \exp\left(-\frac{1}{2l^2}(\mathbf{x}_i - \mathbf{x}_j)^\mathrm{T}(\mathbf{x}_i - \mathbf{x}_j)\right), \qquad (3)$$

where $\sigma_f$ is a hyperparameter that controls the vertical span of the function and $l$ is a hyperparameter that determines the rate at which the correlation between two data points changes as the distance between them increases.

Given a training dataset $(\mathbf{X}, \mathbf{Y})$, for a new data point $\mathbf{x}^*$ the posterior distribution of $y^*$ is Gaussian, i.e.,

$$\mathcal{P}(y^*|\mathbf{x}^*, \mathbf{X}, \mathbf{y}) = \mathcal{N}(\mu^*, \sigma^{*2}), \qquad (4)$$

where $\mu^*$ represents the mean of the predicted distribution and is employed as the point estimate of $y^*$, and the standard deviation $\sigma^*$ of the predicted distribution is used to quantify the uncertainty in the predicted value. The mean of the predicted distribution $\boldsymbol{\mu}^* = \mathbf{K}^{*T}(\mathbf{K} + \sigma_n^2 \mathbf{I})^{-1}\mathbf{y}$ is used as the point estimate of $y^*$, where $\mathbf{K}^*$ is the covariance matrix between $\mathbf{x}^*$ and the data points in the training dataset $\mathbf{X}$, and $\sigma_n^2$ is the variance of independent and identically distributed (i.i.d.) Gaussian noise representing the uncertainty in the training data. The covariance of the predicted distribution is given with $\boldsymbol{\sigma}^{*2} = k(\mathbf{x}^*, \mathbf{x}^*) - \mathbf{K}^{*T}(\mathbf{K} + \sigma_n^2 \mathbf{I})^{-1}\mathbf{K}^*$. GPR is among the most powerful and flexible methods for uncertainty quantification in regression tasks. By using different kernel functions and hyperparameters, GPR allows introducing domain knowledge and adapting the predictive distribution to the specific patterns and trends in a dataset.

## 2.2 Neural Networks with Deterministic Parameters

Neural Networks have been increasingly employed for uncertainty quantification of predicted values. Approaches such as Deep Ensembles and MC Dropout employ standard NNs with deterministic values of the parameters (weight and biases) to obtain probabilistic outputs, allowing for uncertainty quantification. Other works use Bayesian NNs with stochastic parameters for uncertainty quantification.

*2.2.1 Deep Ensemble*

Deep Ensembles (DE) [13] is a conceptually simple approach for generating probabilistic outputs with a collection of standard single-point prediction NNs. DE involves first training multiple NNs for a regression task, and afterward, aggregating their outputs to estimate the prediction uncertainties. The inherent randomness in the initializations of NN parameters and the associated training process, drive the NNs to converge to different solutions in the hypothesis space. As a result, the DE approach results in samples of different network parameters that produce stochastic outputs.

Let's assume an ensemble of $S$ NNs trained on the dataset $(\mathbf{x}_i, y_i) \in \mathcal{D}$ and parameterized with parameters $\theta_i, \theta_2 \ldots, \theta_S$. For a new data point $\mathbf{x}^*$, the DE predictions are treated as a Gaussian distribution, where the predicted mean and standard deviation obtained by averaging the predictions of the ensemble are used as the target value and uncertainties estimates:

$$y^* = \frac{1}{S}\sum_{s=1}^{S} f(\mathbf{x}^*, \theta_s) \ . \tag{5}$$

$$\sigma^* = \sqrt{\frac{1}{S}\sum_{s=1}^{S}\left(y^* - f(\mathbf{x}^*, \theta_s)\right)^2} \ . \tag{6}$$

In general, the individual NNs in the DE can have different architectures, although in most prior works, a NN with the same architecture is trained multiple times. Several related works have also applied bagging (i.e., bootstrap aggregation), where the networks in the DE are trained on random subsets of the training data, to add an additional source of randomness to the training process [41]. The DE approach offers ease of implementation and is suitable for parallelizability and scalability across NNs and datasets. In comparison to other uncertainty quantification methods, a disadvantage is the increased training time and requirement for computational resources, since DE requires to train a group of NNs.

*2.2.2 Monte Carlo Dropout*

In standard NNs, the dropout technique is commonly applied during the training to reduce model complexity and prevent overfitting. It is implemented with a dropout layer that multiplies the output of each neuron by a filter selected with a Bernoulli distribution, randomly turning off (i.e., dropping out) a portion of the neurons. The dropout is not applied during the inference step with standard NNs. Monte Carlo (MC) Dropout [14] is a simple extension of the standard dropout technique, which applies dropout during inference. Consequently, the output from MC Dropout creates a distribution of the predictions by a trained model.

For a new data point $\mathbf{x}^*$, MC Dropout performs $M$ forward passes through the trained network with the dropout enabled to obtain Monte Carlo samples, resulting in $M$ different predictions $f(\mathbf{x}^*, \theta_i)$. Similarly to equations (5) and (6) in the DE approach, the uncertainty estimate is computed from the resulting distribution of predicted target values $f(\mathbf{x}^*, \theta_i)$. The advantages of MC dropout are that it can be used with any network architecture where a dropout layer is applicable, and it provides a computationally inexpensive way for uncertainty estimates with NNs.

## 2.3 Neural Networks with Probabilistic Parameters

Differently from standard NNs with deterministic parameters, Bayesian Neural Networks (BNNs) represent the network parameters with probability distributions, instead of fixed values. BNNs are probabilistic models that allow incorporating prior knowledge into the model and updating the parameters based on observed data. For a BNN model parameterized with parameters $\theta$ that form probability distributions, inference for a new data point $\mathbf{x}^*$ is made by using the posterior predictive distribution $\mathcal{P}(y^*|\mathbf{x}^*, \mathcal{D}) = \int \mathcal{P}(y^*|\mathbf{x}^*, \theta)\, \mathcal{P}(\theta|\mathcal{D}) d\theta$. Direct calculation of the posterior distribution of the parameters given observed data $\mathcal{P}(\theta|\mathcal{D})$ is intractable (and hence, the same applies to the predictive distribution $\mathcal{P}(y^*|\mathbf{x}^*, \mathcal{D})$). Various approximations for $\mathcal{P}(\theta|\mathcal{D})$ have been used in practice, among which the most popular methods are Variational Inference (VI) and Markov Chain Monte Carlo (MCMC).

### 2.3.1 Variational Inference BNN

Variational Inference (VI) BNNs [17] employ an optimization technique to approximate the intractable posterior distribution $\mathcal{P}(\theta|\mathcal{D})$ (that is, $\mathcal{P}(\theta|\mathbf{X}, \mathbf{Y})$) with a simpler parameterized distribution $q_\phi(\theta)$ (referred to as variational distribution) from a family of distributions $\mathcal{Q}$. The VI optimization is as follows:

$$\phi^* = \arg \min_{q_\phi(\theta) \in \mathcal{Q}} D_{\mathrm{KL}}\big[q_\phi(\theta) || \mathcal{P}(\theta|\mathcal{D})\big] \;, \tag{7}$$

where the goal is to calculate the parameters of the variational distribution $q_\phi(\theta)$ that approximates the posterior distribution $\mathcal{P}(\theta|\mathcal{D})$. The Kullback-Leibler (KL) divergence is used as a measure of closeness between the two distributions. Directly calculating the KL divergence is also challenging, since it involves calculating the evidence $\mathcal{P}(\mathbf{Y}|\mathbf{X})$. To address this issue, an alternative approach has been developed, which utilizes the following Evidence Lower Bound (ELBO):

$$ELBO(\phi) = \mathbb{E}_{\theta \sim q_\phi(\theta)}[\log \mathcal{P}(\mathbf{Y}|\mathbf{X}, \theta)] - D_{\mathrm{KL}}\big[q_\phi(\theta) || \mathcal{P}(\theta|\mathcal{D})\big] \;. \tag{8}$$

Since the KL divergence term $D_{\mathrm{KL}}\big[q_\phi(\theta) || \mathcal{P}(\theta|\mathcal{D})\big]$ in (9) is always non-negative, the expected log-likelihood of the data $\log \mathcal{P}(\mathbf{Y}|\mathbf{X}, \theta)$ is always larger than ELBO. Therefore, using a loss function that maximizes ELBO in (8) minimizes the KL divergence between the variational distribution $q_\phi(\theta)$ and the posterior distribution $\mathcal{P}(\theta|\mathcal{D})$.

The predictive uncertainty for new data point $\mathbf{x}^*$ is estimated by sampling from the variational distribution $q_\phi(\theta)$. Using equations (5) and (6), the point estimate and uncertainty are calculated as the mean and standard deviation of the drawn samples from $q_\phi(\theta)$.

### 2.3.2 Markov Chain Monte Carlo (MCMC) BNN

MCMC BNNs [42] approximate the posterior distribution of NN parameters $\theta$ given observational data $\mathcal{P}(\theta|\mathbf{X}, \mathbf{Y})$ through Monte Carlo sampling. The approach employs a Markov Chain of model parameters, where each set of parameters $\theta_i$ is a sample from the posterior distribution. To approximate the posterior distribution, the chain iteratively explores the space of possible parameters $\theta_i$. This exploration is guided by comparing the posterior probabilities, and as the Markov chain evolves, it effectively samples across the entire distribution space, allowing to converge to the target posterior distribution. After reaching a stationary distribution, for new input data point $\mathbf{x}^*$, the set of $S$ generated samples $\{\theta_1, \theta_2, \dots, \theta_S\}$ from $\mathcal{P}(\theta|\mathbf{X}, \mathbf{Y})$ is used to generate $S$ predictions $f(\mathbf{x}^*, \theta_S)$. Point estimates and uncertainty estimates are calculated by averaging the predictions, as in equations (5) and (6).

Several MCMC sampling methods are used for approximating the posterior distribution with BNNs for regression tasks. Metropolis-Hasting algorithm [43] is often employed since it does not require exact knowledge about the probability distribution $\mathcal{P}(\theta)$ to sample from, and a function that is proportional to the distribution is sufficient. Hamiltonian Monte Carlo (HMC) algorithm [44] is a version of Metropolis-Hasting that introduces a momentum term for proposing new states similar to simulating a physical system with Hamiltonian dynamics. Likewise, the No-U-Turn Sampling (NUTS) algorithm [45] is a sub-version of HMS that offers an approach for automatic selection of the hyperparameters.

MCMC method for BNNs is computationally expensive because it requires generating a large number of samples to obtain independent samples from the distribution, and it also requires a large number of iterations for convergence. On the other hand, MCMC BNNs are considered one of the most efficient methods for sampling from the posterior distribution and offer improved results in capturing uncertainty in comparison to VI BNNs, Deep Ensembles, or MC Dropout.

## 3. Proposed Approach

### 3.1 Physics-Informed Machine Learning

Physics-Informed Machine Learning (PIML) is an approach designed to integrate insights of the fundamental physics laws governing a process into ML models, in order to enhance the consistency of the predictions [29-33][46][47]. This approach differs from the broadly used supervised ML methods, and instead of relying solely on observational data points to model a process, PIML leverages existing physics knowledge of a specific task to guide the model solutions. Specifically, prior knowledge about a task is employed to restrict the hypothesis space in ML models to architectural and other choices, introducing inductive biases. PIML models incorporate additional learning biases based on the fundamental physical rules guiding the process, commonly implemented as mathematical constraints in the learning algorithm. Such constraints are either encoded into the loss function of a PIML model (referred to as physics-informed loss function), can be enforced by designing custom physics-informed layers in the network architecture, or can be applied via physics-informed feature selection or feature engineering from the original high-dimensional input data. Similarly, prior physics knowledge can be introduced into PIML models in the form of differential or integral equations. These flexibilities in designing PIML models provided by combining different types of mathematical formulations, prior task knowledge, and physical constraints are beneficial in guiding the models toward solutions that are more accurate and adhere to the governing physics laws of the process.

PIML can potentially address challenges associated with modeling material properties, as it leverages the demonstrated capability of ML methods—especially deep neural networks—to capture intricate relationships within high-dimensional multi-scale and multi-physics data. Namely, accurately representing the dynamics and deformation mechanisms of materials with physics-based models (e.g., via partial differential equations) poses insurmountable difficulties, since it is exceptionally challenging to mathematically define all different underlying processes that change over time. Indeed, existing physics-based models capture only the most important factors that influence material properties, and are missing fine details of the underlying physics. PIML also can address the challenges posed by the limited availability of large curated datasets in materials science. Therefore, the fusion of historical experimentally collected material property data and physics-based models within a PIML framework holds the potential to enhance long-term predictions of material behavior under different conditions and exceeds the capabilities of traditional physics-based models.

In our proposed approach, we developed a PIML framework for predicting creep rupture life in metal alloys by introducing physics-informed feature engineering to augment the set of input features to the regression models and by applying a physics-informed loss function that introduces physics constraints into the learning algorithm.

*3.1.1 Physics-Informed Feature Engineering*

We introduce two categories of physics-informed features based on estimations of the creep rupture life and stacking fault energy by using physics-based models.

*3.1.1.1 Creep Rupture Life Estimation*

Creep is a slow irreversible deformation process under the influence of stresses below the yield stress of a material. The prediction of creep rupture life, related to the time duration that a material can sustain before undergoing rupture is essential for guiding design strategies, maintenance regimens, and safety protocols for systems and structures. Existing physics-based creep models are broadly classified into two major categories: time-temperature parametric (TTP) models and creep constitutive (CC) models.

TTP models derive equations of thermal creep in metal materials by assuming interdependence between the effects of time and temperature on creep rupture life. The relationship between creep rupture life $t_f$, temperature $T$, and stress $\sigma$ is established as $\mathcal{P}(t_f, T) = f(\sigma)$, where the function $\mathcal{P}$ combines the rupture time and temperature into a single parameter. Well-known TTP models include the Larson-Miller method [48] $\mathcal{P}(t_f, T) = T \cdot (C_{\text{LM}} + \log t_f)$, Manson-Haferd method [49] $\mathcal{P}(t_f, T) = (\log t_f - \log t_{\text{in}})/(T - T_{\text{in}})$, and Orr-Sherby-Dorn method [50] $\mathcal{P}(t_f, T) = \log t_f - (Q_C/2.3RT)$. In these formulations, $C_{\text{LM}}$ is a constant, $Q_C$ is the creep activation energy, $R$ denotes the universal gas constant, and $t_{\text{in}}, T_{\text{in}}$ are constants that represent the point of intersection of the iso-stress lines in $\log t_f$ versus $T$ plots. The stress function $f(\sigma)$ is typically represented as a cubic polynomial logarithmic function $(\sigma) = c_0 + c_1 \log \sigma + c_2 \log^2 \sigma + c_3 \log^3 \sigma$, where the coefficients $c_0, c_1, c_2, c_3$ are obtained via least-square regression fit to short-term experimental data. For a given material, based on establishing the relationship between the stress $f(\sigma)$ versus the time-temperature parameter $\mathcal{P}(t_f, T)$ from available short-term creep measurements, TTP methods extrapolate the plots to longer times to estimate the creep rupture life $t_f$.

Creep constitutive (CC) models are based on the Monkman-Grant (MG) conjecture [51], which postulates that the product of the creep rupture life $t_f$ and exponentiated minimum creep strain rate $\dot{\varepsilon}_{\min}$ for isothermal conditions is constant. That is, $t_f \cdot \dot{\varepsilon}_{\min}^n = C_{\text{MG}}$, where $n$ is the creep strain rate exponent, and $C_{MG}$ is a constant. Researchers have proposed several CC models to enhance the creep life predictions. For instance, an alternative formulation of the creep strain rate is $\dot{\varepsilon} = ae^{-\frac{Q_{\text{in}}}{RT}} \cdot \sinh(be^{-\frac{Q_{\text{pw}}}{RT}} \sigma)$ [52], where the first term $ae^{-\frac{Q_{\text{in}}}{RT}}$ describes the power-law creep mechanism in the high-stress range, and the term $\sinh(be^{-\frac{Q_{\text{pw}}}{RT}} \sigma)$ describes the viscous creep under the diffusion mechanism in moderate and low-stress ranges. The coefficients $a$ and $b$, and the activation energies $Q_{\text{in}}$ and $Q_{\text{pw}}$ are obtained by fitting to experimental data of creep strain rate $\dot{\varepsilon}$ for given temperature $T$ and stress $\sigma$. With the progress in material science, we can expect the development of novel models that more accurately predict creep behavior.

In this work, we employed the Manson-Haferd method for estimating the creep rupture life. The motivation is because the Manson-Haferd method was used for modeling the creep rupture life in SS316 alloys in the NIMS database, and the values of the coefficients $c_0, c_1, c_2, c_3$ for the least-square regression fit are provided in the database. Accordingly, we used the Manson-Haferd method for estimating the creep rupture life for the Nickel-based superalloys and Titanium alloys datasets.

*3.1.1.2 Stacking Fault Energy*

Stacking fault energy (SFE) represents the energy difference between atoms within the regular lattice structure and those located in the stacking fault region. It defines how resistant a material is to deformation occurring along specific crystallographic planes. SFE is an important parameter that impacts the strength and deformation behavior of steel materials. To calculate SFE of stainless steels $\gamma_{SFE}$, we used the following equation [53]

$$\gamma_{SFE} = \gamma^0 + 1.59Ni - 1.34Mn + 0.06Mn^2 - 1.75Cr + 0.01Cr^2 + 15.21Mo - 5.59Si - 60.69(C + 1.2N)^{1/2} + 26.27(C + 1.2N) * (C + 1.2Cr + Mn + Mo)^{1/2} + 0.6[Ni * (Cr + Mn)]^{1/2} , \qquad (9)$$

where SFE is approximated as a function of the chemical composition of the material where Ni, Mn, Cr, Mo, Si, C, and N denote weight percentages of the alloying elements, and $\gamma^0$ is a constant equal to $39 \, mJ/m^2$ representing the SFE of pure austenitic iron at room temperature.

*3.1.2 Physics-Informed Loss Function*

The PIML paradigm allows introducing initial, boundary conditions, and other types of physics constraints into the loss function of a learning algorithm. Motivated by the work of Zhang et al. [31], we introduced two physics-informed boundary constraints regarding the predicted creep rupture life into the loss of NNs. The first introduced loss term $\mathcal{L}_{PI-B1} = \frac{1}{N}\sum_{i=1}^{N} ReLU(-y_i^*)$ imposes that the predicted creep rupture life by the model $y^*$ is non-negative. The second introduced loss term $\mathcal{L}_{PI-B2} = \frac{1}{N}\sum_{i=1}^{N} ReLU(y_i^* - a)$ enforces that the predicted creep rupture life is upper bounded by a constant $a$. For the constant $a$ we adopted the value of 100,000 hours, because it is the greatest creep rupture life value in the three datasets. In these loss terms, ReLU denotes Rectified Linear Unit activation function, defined as $ReLU(x) = \{0 \text{ for } x < 0, \; x \text{ for } x \geq 0\}$. The two physics-informed terms regard negative and excessively large creep life values as physical violations that should be prevented from being output by the model. The two terms are added to the standard mean-squared error loss for regression tasks $\mathcal{L}_{NN} = \frac{1}{N}\sum_{i=1}^{N}(y - y_i^*)^2$, resulting in a composite physics-informed loss function: $\mathcal{L} = \mathcal{L}_{NN} + \lambda_1 \mathcal{L}_{PI-B1} + \lambda_2 \mathcal{L}_{PI-B2}$, where $\lambda_1$ and $\lambda_2$ are weighting coefficients, which are empirically determined to quantify the contributions of the terms $\mathcal{L}_{PI-B1}$ and $\mathcal{L}_{PI-B2}$, respectively.

## 3.2 Application Case: Active Learning

Supervised ML requires that for each input data, there is an associated target data point. Generally, for a supervised ML model to perform well, it often needs to be trained with a large number of labeled data points. In reality, labeled data may be scarce and expensive to obtain since the labeling or annotation process is time- and cost-consuming, whereas unlabeled data could often be accessed easily. Additionally, among the labeled data points, some could carry similar information, thus, contributing less value compared to the data points that carry dissimilar information. The Active Learning (AL) method was developed to select the most informative data points that speed up the training process [19].

In pool-based AL, the training dataset $\mathcal{D}$ comprises a pool of unlabeled data $\mathcal{D}^U$ and labeled data $\mathcal{D}^L$. An initial ML model $f$ is first trained with a small, randomly selected labeled dataset $C \in \mathcal{D}^L$. Next, a query strategy is applied to select the most informative data $\mathcal{D}^A \in \mathcal{D}^U$ by employing an acquisition function to measure and rank the informativeness of the data. An annotator is asked to label the data $\mathcal{D}^A$ selected by the query strategy, which is added to the dataset $C$, and the model is re-trained with the updated dataset. These steps are iteratively repeated until the model $f$ converges [36].

Two main types of query strategies are: uncertainty/informativeness-based strategies that ensure the informativeness of the unlabeled data, and representative-based/diversity-based strategies that measure the similarity of the instances and deal with issues such as sampling bias and inclusion of outliers. In addition, a hybrid query strategy combining these two strategies can also be applied [36].

In this paper, we apply a hybrid query strategy that includes a Variance Reduction (VR) uncertainty-based method, and a *k*-means clustering diversity-based batch mode method to guide the selection of the newly added data [9].

Variance reduction (VR) has been proven to be an effective AL method with regression tasks [54], where the goal is to minimize the variance of the model. For a hypothesis $f^{(C^{(i)})}$ learned on $C$, and a true hypothesis $f^*$, the total expectation of the error $E_{out}(f^{(J^{(i)})}) = E_\mathbf{x}[(f^{(C^{(i)})}(\mathbf{x}) - f^*(\mathbf{x}))]^2$, and the average estimation of learned hypotheses is:

$$\bar{f} \approx \frac{1}{K}\sum_{k=1}^{K}\left(f^{(C^{(i)})}\right), \tag{10}$$

where $k$ is a data point in $C$. The expectation on the generalization error of the entire dataset is:

$$E_D[E_{out}(f^{(C')})] = E_\mathbf{x}[Variance(\mathbf{x}) + bias(\mathbf{x})]. \tag{11}$$

Equation (11) indicates that for a given model, if the bias of the model is fixed, minimizing the variance of the model results in minimal generalization [55]. Therefore, VR selects and annotates the samples with the highest prediction variances, i.e., for which the model is most uncertain of inferring with these samples. Adding these samples to the training dataset reduces the overall generalization error of the model [36].

Conventional AL queries a single data point at each iteration, which is inefficient and leads to iterative training with small changes to the training dataset. To avoid these issues, we implemented Batch Mode Active Learning (BMAL) [56]. BMAL selects and annotates multiple samples $B = \{B^{(1)}, B^{(2)}, ..., B^{(m)}\} \in \mathcal{D}^U$ in each iteration and add these samples to the training dataset to re-train the model. On the other hand, an uncertainty-based BMAL query strategy may not be ideal since the most uncertain samples may be similar to each other. Therefore, diversity-based query strategy is typically preferred in BMAL. Clustering methods are commonly used to group the unlabeled dataset by similarity, and the most dissimilar samples are added to the training dataset. In this paper, we used *k*-means clustering to group the unlabeled samples into $K$ clusters, where each group will have the least feature correlation with each other, and in each cluster the sample with the highest variance will be selected. Therefore, $K$ number of samples will be selected and annotated for training.

## 4. Experiments

### 4.1 Datasets

We used three datasets to conduct experimental validation of the UQ frameworks for creep rupture life prediction.

The first creep dataset of Stainless Stee (SS) 316 alloys was obtained from the National Institute for Materials Science (NIMS) database [57]. The dataset contains 617 test samples with 20 features per sample. Specifically, the features provide information about the material composition related to the mass percent of the elements C, Si, Mn, P, S, Ni, Cr, Mo, Cu, Ti, Al, B, N, Nb+Ta and the material group of the alloy (there are 20 material groups in total), testing conditions including the applied stress (MPa) and temperature (°Celsius), test measurements related to the percentage of elongation and percentage of area reduction, and the recorded creep rupture life (hours).

The second creep dataset is for Nickel-based superalloys and was adopted from the work by Han et al. [58]. The dataset includes 153 test samples with 15 features per sample. The features include material composition related to the weight percentage of the elements Ni, Al, Co, Cr, Mo, Re, Ru, Ta, W, Ti, Nb, and T, testing conditions including applied stress (MPa) and temperature (°Celsius), and the recorded creep rupture life (hours).

The third creep dataset pertains to Titanium alloys and was adopted from Swetlana et al. [59]. It consists of 177 test samples with 24 features per sample. The features provide information about the weight percentage of the elements Ti, Al, V, Fe, C, H, O, Sn, Mb, Mo, Zr, Si, B, and Cr, testing conditions including applied stress (MPa) and temperature (°Celsius), finishing conditions related to the solution treated temperature (°Celsius), solution treated time (hours), annealing temperature (°Celsius), annealing time (hours), test measurements of the steady-state strain rate (1/seconds) and strain to rupture (%), and the recorded creep rupture life (hours).

**4.2 Evaluation Metrics**

The next section describes the used set of metrics to evaluate the predictive accuracy of the models on unseen samples, and the quality of uncertainty quantification.

*4.2.1 Predictive Accuracy Metrics*

*Pearson Correlation Coefficient* (PCC) quantifies the magnitude and direction of the linear relationship between two continuous variables. For a target variable $\mathbf{Y}$ and predicted values by a model $\mathbf{Y}^*$, PCC is calculated as

$$PCC_{\mathbf{Y},\mathbf{Y}^*} = \frac{\text{cov}(\mathbf{Y},\mathbf{Y}^*)}{\sigma_{\mathbf{Y}}\sigma_{\mathbf{Y}^*}}, \tag{12}$$

where $\text{cov}(\mathbf{Y}, \mathbf{Y}^*)$ denotes the covariance and $\sigma_{\mathbf{Y}}$, $\sigma_{\mathbf{Y}^*}$ are the standard deviations of $\mathbf{Y}$ and $\mathbf{Y}^*$. PCC ranges from -1 to +1, where a larger PCC implies a higher degree of linear correlation between the target and predicted variables.

*Coefficient of determination* ($R^2$) quantifies the proportion of the variance in the target variable $\mathbf{Y}$ that is explained by the predicted values from the regression model $\mathbf{Y}^*$. It is defined as

$$R^2 = 1 - \frac{\sum(y_i - y_i^*)^2}{\sum(y_i - \bar{y})^2}, \tag{13}$$

where the numerator represents the sum of squared residuals, and the denominator calculates the sum of squares between the target values and the mean of $\mathbf{Y}$, denoted $\bar{y}$. The coefficient of determination measures the goodness of fit of a regression model and ranges from 0 to 1. Higher $R^2$ values indicate that a larger proportion of the variance in the target variable samples is explained by the predicted values by the model.

*Root-mean-square error* (RMSE) measures the average difference between the target values and the predicted values by a regression model. It is calculated as the standard deviation of the residuals:

$$\text{RMSE} = \sqrt{\frac{\sum_{i=1}^{N}(y_i - y_i^*)^2}{N}} \quad . \tag{14}$$

*Mean absolute error* (MAE) measures the average magnitude of the errors in a set of predicted values, obtained as

$$\text{MAE} = \frac{\sum_{i=1}^{N}|y_i - y_i^*|}{N} \quad . \tag{15}$$

Smaller values of RMSE and MAE indicate better predictive accuracy of a regression model.

*4.2.2 Uncertainty Quantification Metrics*

*Coverage* is a metric that quantifies the proportion of target values **Y** that fall within the predicted uncertainty interval by a regression model. A high coverage implies accurate uncertainty quantification by the model. Values of the coverage metric that are close to the nominal confidence interval of 95% are preferred for reliable uncertainty quantification. In some statistical works, the coverage metric is referred to as validity, since it assesses whether the predicted uncertainty intervals are valid.

*Mean interval width* calculates the average size of the predicted interval around the point estimates, related to the upper and lower uncertainty bounds. I.e., this metric assesses how tight the uncertainty bounds are across all predicted values. Smaller values of the interval width are preferred, as they indicate more precise uncertainty estimates by a regression model. In some statistical works, this metric is also referred to as sharpness.

*Composite metric* was introduced here, in order to capture the trade-offs between the coverage and mean interval width metrics. Specifically, since the coverage and mean interval width metrics alone do not quantify well the accuracy of uncertainty estimation, we introduced the composite metric as: $0.75 \cdot \text{Coverage} + \frac{0.25}{\text{Mean Interval Width}}$. We selected the values 0.75 and 025 for the weight coefficients of the coverage and mean interval width based on empirical evaluation. It is our hope that this metric can combine the impact of the coverage and mean interval width on the uncertainty estimates, where greater values of the composite metric indicate more reliable uncertainty quantification.

**4.3 Implementation Details**

For the regression models, we used the creep rupture life as a target variable, and the remaining features in the datasets were used as inputs to the models. Preprocessing of the input features involved normalization of the values in the range between 0 and 1. Similarly to related works, we applied base 10 logarithm transformation to the values of the creep rupture life. For evaluation of the performance of the different models, we used 5-fold cross-validation.

For the implementation of QR, NGBoost, and GPR, we adopted the same hyperparameters as in the work by Mamun et al. [9]. Following this work, QR used CatBoost model for calculating the quantiles, and we applied the same kernel functions for GPR as in Mamun et al. [9]. We used the scikit-learn, ngboost, and catboost libraries for model training and evaluation.

For implementing the BNN-VI method, the architecture consists of two fully-connected layers with 100 neurons in each layer. We applied Rectified Linear Unit (ReLU) activation function to the output of each layer. For the prior distribution of the network parameters, we adopted a normal distribution with a mean of 0 and a standard deviation of 0.06. The loss function is based on equation (8), with the weight coefficient for the KL divergence term set to 0.01. We used a Stochastic Gradient Descent (SGD) optimizer with Nesterov Momentum set to 0.95, and the learning rate was set to 0.001. After training the model, for inference we generated 1,000 samples from the variational distribution. The point estimates and uncertainty estimates are calculated as the mean and $\pm 3$ standard deviations of the drawn samples, according to (5) and (6).

For the BNN-MCMC approach, we selected an architecture with three fully-connected layers with 10 neurons in each layer. For the prior distribution of the network parameters, we adopted a normal distribution with a mean of 0 and a standard deviation of 1. For approximating the posterior distribution we used the No-U-Turn Sampling (NUTS) algorithm for training the model. For inference, we drew 100 samples, and the point estimates and uncertainty estimates were calculated similarly to the BNN-VI methods by taking the arithmetic mean and $\pm 3$ standard deviations of the generated samples.

For the Deep Ensemble method, we used an ensemble of 5 base learners, each of which is a standard NN with three fully-connected layers having 10 neurons in each layer. Each hidden layer is followed with a dropout layer with a rate of 0.5 and a ReLU activation layer. Mean-square error (MSE) loss function was used, and the Adaptive Moment Estimation (Adam) optimizer with a learning rate of 0.01 was selected for training the models. The final predictions were calculated by taking the arithmetic mean and $\pm 3$ standard deviations from the outputs of the base learners.

For the MC Dropout approach, we used an NN with three fully-connected layers having 100 neurons in each layer. Similar to the Deep Ensemble, we used a dropout rate of 0.5, ReLU activation function, MSE loss, and Adaptive Gradient Algorithm (Adagrad) optimizer with a learning rate of 0.01. For inference, we generated 1,000 predictions, which were used to calculate the point estimates and uncertainty estimates.

For comparison, we implement standard NNs with deterministic parameters, consisting of three fully-connected layers with 1000, 200, and 40 neurons, respectively. We used ReLU activation function after each hidden layer. The loss function was MSE, and the Root Mean Squared Propagation (RMSprop) optimizer was used with a learning rate of 0.01.

BNN-MCMC, BINN-VI, Deep Ensemble, and MC Dropout were implemented using the PyTorch library. For building BNN-MCMC we used the Pyro library [60], and BNN-VI was implemented with the torchbnn library [61]. For the above models, we chose a batch size of 16 data points. The link to our codes is provided in the Abstract section.

### 4.4 Experimental Results

Experimental validation includes a comparative analysis of UQ with the three datasets by using the above-described eight methods. The results are presented in section 4.4.1. For the top four performing approaches for UQ, we evaluated the proposed PIML framework, with the results presented in section 4.4.2. In section 4.4.3, we present the results from the AL study for the GPR, BNN-VI, and BNN-MCMC models.

*4.4.1 Uncertainty Quantification*

For the SS316 alloys dataset, Table 1 presents the average and the standard deviation (in the subscript) of the metrics for predictive accuracy and uncertainty estimations for the considered eight models based on five-fold cross-validation. High values of PCC indicate high correlations between the model predictions and the experimentally measured creep rupture life, and high values of $R^2$ point to more consistent fit of the predicted values to the experimental creep rupture life. Similarly, low values of RMSE and MAE imply that the predicted creep rupture life more closely aligns with the experimentally measured creep rupture life. The results show that the NN model and Bayesian NNs performed better than the traditional ML models.

The BNN-MCMC approach achieved the best performance for all point accuracy metrics, including PCC, $R^2$, RMSE, and MAE. In addition, BNN-MCMC produced the best results for the Interval width and Composite Metric, except for the Coverage for which BNN-VI had the highest coverage. As expected, GPR achieved comparable performance and it was the second best-performing method.

Figure 1 shows the target creep life, predicted creep life, and uncertainty estimates by the eight methods on one fold of the test dataset. One can note that the uncertainty estimates by QR and NGBoost are over-estimated. On the other hand, GPR provides accurate uncertainty estimates, as well as, the deterministic NNs and BNNs have generally good performance for point predictions and uncertainty estimates.

*Table 1. Experimental results for the SS316 alloys dataset, including predictive accuracy metrics (PCC, $R^2$, RMSE, MAE) and uncertainty quantification metrics (coverage, interval width, composite metrics) for 8 compared methods: NN (Neural Network), QR (Quantile Regression), NGBoost (Natural Gradient Boosting), GPR (Gaussian Process Regression), Deep Ensemble, MC Dropout, BNN – VI (Variational Inference), and BNN – MCMC (Markov Chain Monte Carlo).*

|  | PCC ↑ | $R^2$ ↑ | RMSE ↓ | MAE ↓ | Coverage ↑ | Interval width ↓ | Composite metric ↑ |
|---|---|---|---|---|---|---|---|
| **NN** | 0.988±0.004 | 0.973±0.011 | 0.145±0.031 | 1.027±0.039 |  |  |  |
| **QR** | 0.958±0.011 | 0.913±0.022 | 0.260±0.028 | 0.195±0.023 | 86.23±3.13 | 1.98±0.22 | 0.77±0.02 |
| **NGBoost** | 0.917±0.025 | 0.826±0.045 | 0.368±0.042 | 0.13±0.037 | 92.39±1.87 | 1.34±0.18 | 0.88±0.02 |
| **GPR** | 0.993±0.001 | 0.987±0.002 | 0.102±0.007 | 0.072±0.003 | 94.33±2.54 | 0.39±0.01 | 1.44±0.03 |
| **Deep Ensemble** | 0.991±0.003 | 0.983±0.006 | 0.115±0.021 | 0.084±0.018 | 87.20±3.92 | 0.39±0.08 | 1.36±0.13 |
| **MC Dropout** | 0.988±0.002 | 0.977±0.003 | 0.134±0.012 | 0.101±0.007 | 77.00±3.69 | 0.33±0.04 | 1.38±0.07 |
| **BNN - VI** | 0.984±0.006 | 0.962±0.014 | 0.172±0.030 | 0.130±0.017 | **95.46**±2.49 | 0.69±0.04 | 1.09±0.03 |
| **BNN - MCMC** | **0.996**±0.001 | **0.991**±0.001 | **0.085**±0.008 | **0.060**±0.004 | 92.71±3.61 | **0.30**±0.03 | **1.57**±0.07 |

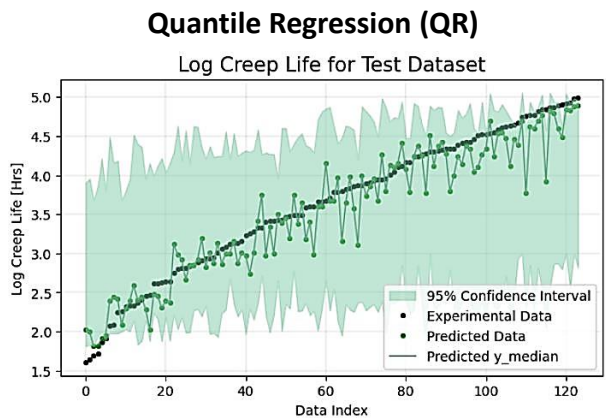
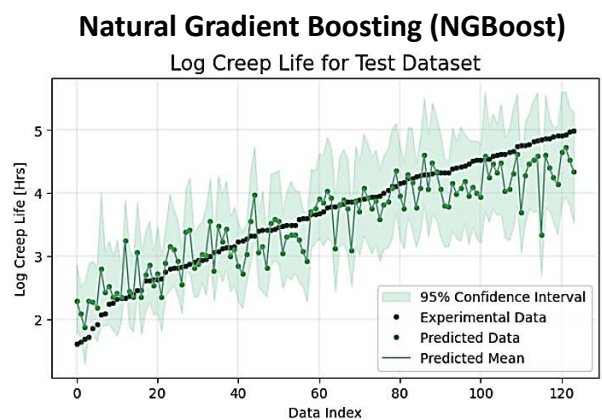

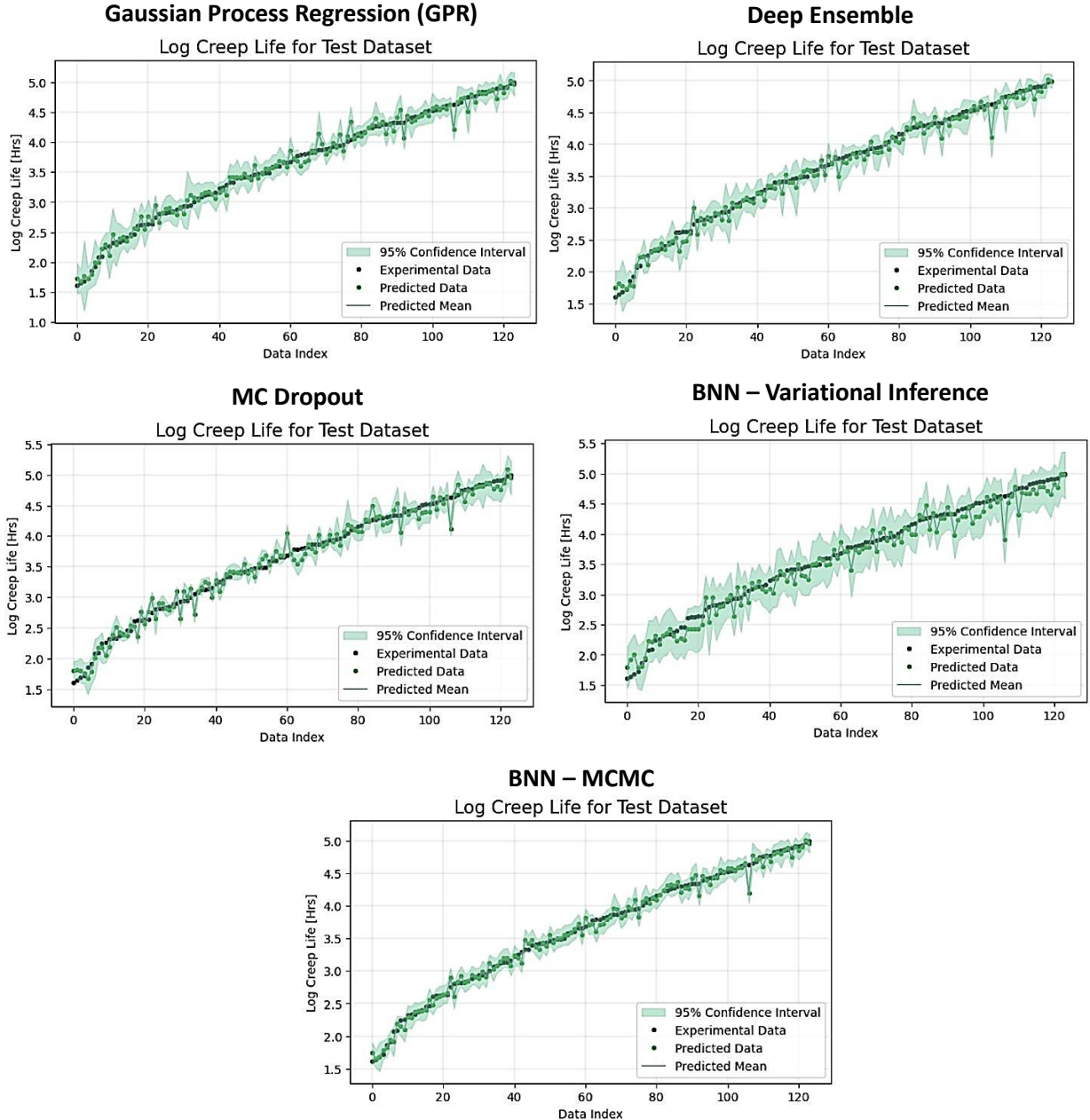

*Figure 1.* Experimental data points, predicted data points, and uncertainty estimates for Quantile regression (QR), Natural Gradient Boosting (NGBoost), Gaussian Process Regression (GPR), Deep Ensemble, MC Dropout, BNN-Variational Inference, and BNN-Markov Chain Monte Carlo (MCMC) for the SS316 alloys dataset. The logarithm values for the creep rupture life are shown on a logarithmic axis. The green shaded area represents the 95% confidence interval for the predictions by the models.

The corresponding results for the Nickel-based superalloy dataset are shown in Table 2. Since this dataset is of smaller size and has only 153 data points, there is a significant performance drop for most models. We can still note that BNN-MCMC, BNN-VI, and GPR have comparable predictions, and BNN-MCMC has a small advantage in point estimate predictions. The values of the uncertainty estimates indicate greater variability due to the challenges associated with smaller datasets. The results for predicted creep rupture life and uncertainty estimates are shown in Figure 2 only for the three best-performing models: GPR, BNN-VI, and BNN-MCMC.

*Table 2. Experimental results for the Nickel-based superalloys dataset, including predictive accuracy metrics (PCC, $R^2$, RMSE, MAE) and uncertainty quantification metrics (coverage, interval width, composite metrics) for 8 compares methods: NN (Neural Network), QR (Quantile Regression), NGBoost (Natural Gradient Boosting), GPR (Gaussian Process Regression), Deep Ensemble, MC Dropout, BNN – VI (Variational Inference), and BNN – MCMC (Markov Chain Monte Carlo).*

|  | PCC ↑ | $R^2$ ↑ | RMSE ↓ | MAE ↓ | Coverage ↑ | Interval width ↓ | Composite metric ↑ |
|---|---|---|---|---|---|---|---|
| **NN** | 0.806$_{\pm 0.113}$ | 0.620$_{\pm 0.190}$ | 0.241$_{\pm 0.030}$ | 0.427$_{\pm 0.060}$ |  |  |  |
| **QR** | 0.819$_{\pm 0.048}$ | 0.641$_{\pm 0.096}$ | 0.243$_{\pm 0.032}$ | 0.191$_{\pm 0.024}$ | 83.69$_{\pm 8.66}$ | 0.997$_{\pm 0.148}$ | 0.895$_{\pm 0.030}$ |
| **NGBoost** | 0.745$_{\pm 0.101}$ | 0.554$_{\pm 0.154}$ | 0.267$_{\pm 0.035}$ | 0.213$_{\pm 0.030}$ | **95.41$_{\pm 2.60}$** | 1.044$_{\pm 0.170}$ | 0.967$_{\pm 0.044}$ |
| **GPR** | 0.907$_{\pm 0.019}$ | 0.801$_{\pm 0.054}$ | 0.175$_{\pm 0.026}$ | 0.125$_{\pm 0.011}$ | 92.77$_{\pm 5.56}$ | 0.594$_{\pm 0.019}$ | 1.123$_{\pm 0.048}$ |
| **Deep Ensemble** | 0.875$_{\pm 0.074}$ | 0.748$_{\pm 0.126}$ | 0.196$_{\pm 0.032}$ | 0.143$_{\pm 0.028}$ | 69.27$_{\pm 10.6}$ | 0.414$_{\pm 0.098}$ | **1.454$_{\pm 0.119}$** |
| **MC Dropout** | 0.858$_{\pm 0.075}$ | 0.717$_{\pm 0.149}$ | 0.207$_{\pm 0.046}$ | 0.148$_{\pm 0.027}$ | 45.72$_{\pm 7.43}$ | **0.209$_{\pm 0.017}$** | 0.654$_{\pm 0.094}$ |
| **BNN - VI** | 0.909$_{\pm 0.032}$ | 0.791$_{\pm 0.069}$ | 0.185$_{\pm 0.030}$ | 0.147$_{\pm 0.025}$ | 87.52$_{\pm 5.80}$ | 0.534$_{\pm 0.029}$ | 1.148$_{\pm 0.044}$ |
| **BNN - MCMC** | **0.914$_{\pm 0.027}$** | **0.824$_{\pm 0.050}$** | **0.167$_{\pm 0.023}$** | **0.116$_{\pm 0.019}$** | 92.84$_{\pm 4.71}$ | 0.607$_{\pm 0.061}$ | 1.122$_{\pm 0.021}$ |

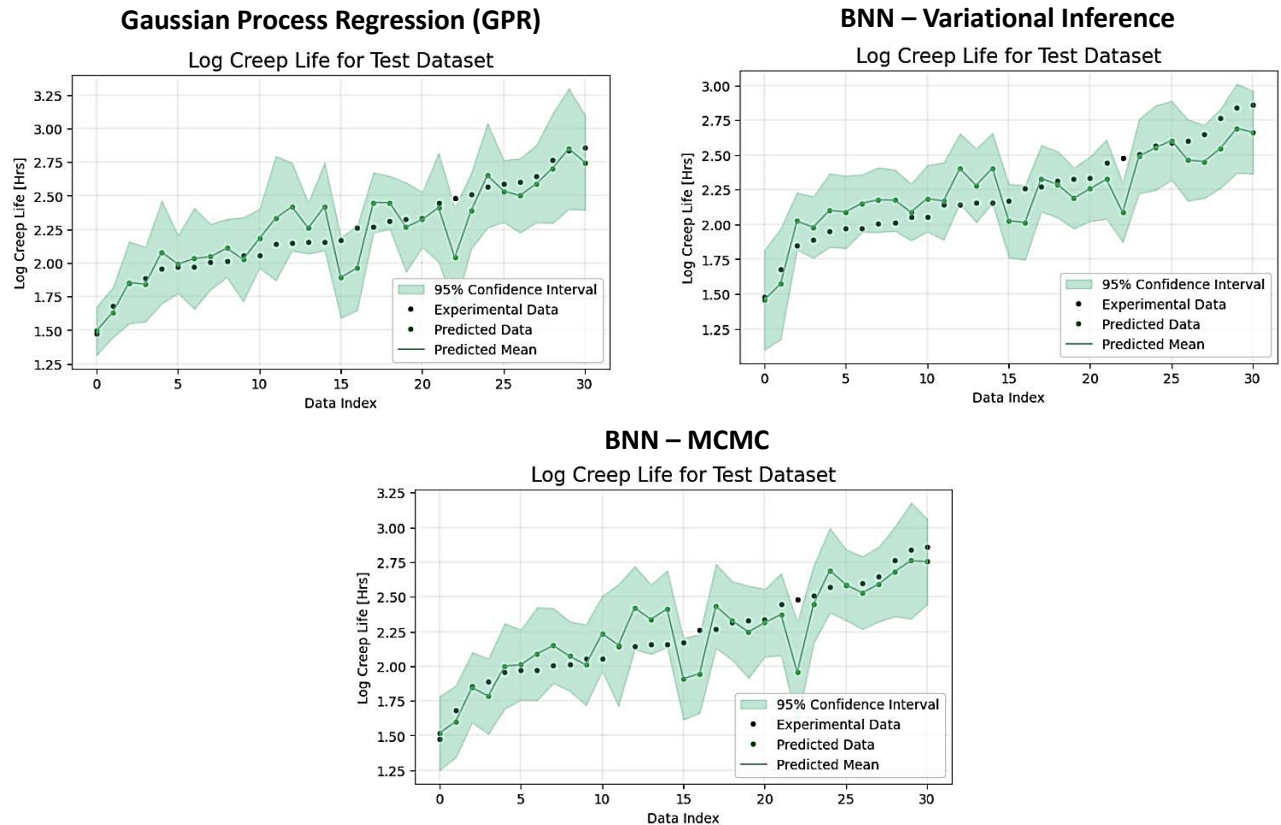

*Figure 2. Experimental data points, predicted data points, and uncertainty estimates for Gaussian Process Regression (GPR), BNN-Variational Inference, and BNN-Markov Chain Monte Carlo (MCMC) for the Ni-based superalloys dataset.*

*The logarithm values for the creep rupture life are shown on a logarithmic axis. The green shaded area represents the 95% confidence interval for the predictions by the models.*

Next, we evaluated the models on the Titanium alloy dataset, which similar to the Nickel superalloys dataset is also much smaller than the SS316 dataset. The overall performance is consistent with the results in Tables 1 and 2, with the top performers being BNN-MCMC and GPR. The mean interval widths are wider for the top performers, indicating that the models are less confident about the uncertainty predictions. Figure 3 presents the predicted creep rupture life and uncertainty estimates for the three best-performing models: GPR, BNN-VI, and BNN-MCMC.

*Table 3. Experimental results for the Titanium alloys dataset, including predictive accuracy metrics (PCC, $R^2$, RMSE, MAE) and uncertainty quantification metrics (coverage, interval width, composite metrics) for 8 compares methods: NN (Neural Network), QR (Quantile Regression), NGBoost (Natural Gradient Boosting), GPR (Gaussian Process Regression), Deep Ensemble, MC Dropout, BNN – VI (Variational Inference), and BNN – MCMC (Markov Chain Monte Carlo).*

|  | PCC ↑ | $R^2$ ↑ | RMSE ↓ | MAE ↓ | Coverage ↑ | Interval width ↓ | Composite Metric ↑ |
|---|---|---|---|---|---|---|---|
| **NN** | 0.794±0.227 | 0.381±0.777 | 0.800±0.481 | 1.326±0.150 | | | |
| **QR** | 0.584±0.263 | 0.348±0.320 | 0.396±0.350 | 1.33±0.099 | 81.98±6.67 | **0.43**±0.19 | 0.616±0.050 |
| **NGBoost** | 0.510±0.208 | 0.39±0.395 | 0.453±0.277 | 1.92±0.073 | 94.90±2.15 | 0.24±0.75 | 0.712±0.016 |
| **GPR** | 0.921±0.034 | 0.839±0.066 | 0.471±0.081 | **0.309**±0.044 | 93.24±2.22 | 1.99±0.20 | 0.866±0.017 |
| **Deep Ensemble** | 0.903±0.054 | 0.800±0.109 | 0.517±0.106 | 0.333±0.041 | 79.10±1.37 | 1.11±0.171 | **0.965**±0.057 |
| **MC Dropout** | 0.883±0.101 | 0.641±0.423 | 0.623±0.345 | 0.348±0.110 | 54.84±5.62 | 0.51±0.05 | 0.915±0.049 |
| **BNN - VI** | 0.919±0.039 | 0.811±0.116 | 0.474±0.096 | 0.320±0.040 | 75.10±9.36 | 0.90±0.06 | 0.862±0.091 |
| **BNN - MCMC** | **0.922**±0.029 | **0.843**±0.061 | **0.449**±0.050 | 0.336±0.023 | **94.87**±4.21 | 1.77±0.14 | 0.860±0.026 |

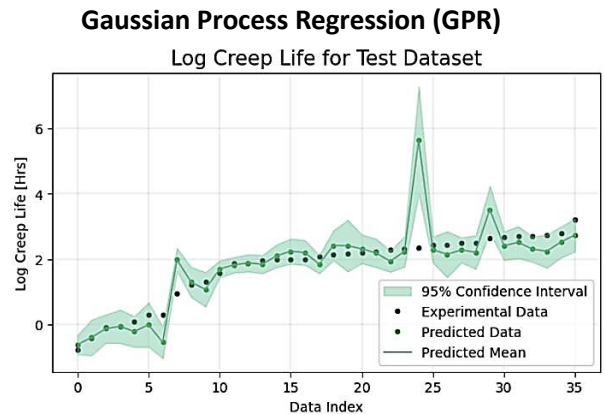
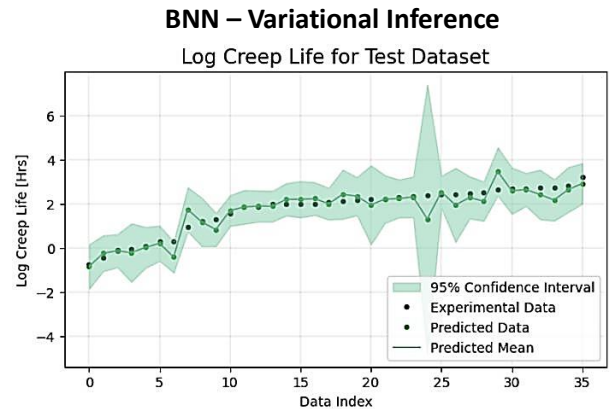

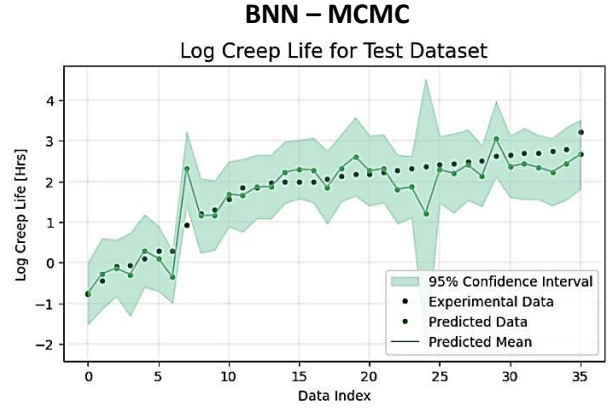

*Figure 3. Experimental data points, predicted data points, and uncertainty estimates for Gaussian Process Regression (GPR), BNN-Variational Inference, and BNN-Markov Chain Monte Carlo (MCMC) for the Ti-based alloys dataset. The logarithm values for the creep rupture life are shown on a logarithmic axis. The green shaded area represents the 95% confidence interval for the predictions by the models.*

#### 4.4.2 Physics-Informed Machine Learning

Experimental validation encompasses the NNs and three best-performing models from the previous section: GPR, BNN-VI, and BNN-MCMC. We introduced physics-informed features and a physics-informed loss function for the NN method. For the SS316 alloys dataset, the results are presented in Table 4. We can notice that the PIML framework improves the points estimates and uncertainty estimates for most regression models. Only for the PI-BNN-MCMC, the performance slightly decreased in comparison to the BNN-MCMC model without physics knowledge.

*Table 4. Experimental results for the SS316 alloys dataset with physics-informed features, including predictive accuracy metrics (PCC, $R^2$, RMSE, MAE) and uncertainty quantification metrics (coverage, interval width, composite metrics) for 8 compares methods: NN (Neural Network), QR (Quantile Regression), NGBoost (Natural Gradient Boosting), GPR (Gaussian Process Regression), Deep Ensemble, MC Dropout, BNN – VI (Variational Inference), and BNN – MCMC (Markov Chain Monte Carlo).*

|  | PCC ↑ | $R^2$ ↑ | RMSE ↓ | MAE ↓ | Coverage ↑ | Interval width ↓ | Composite Metric ↑ |
|---|---|---|---|---|---|---|---|
| **NN** | $0.988_{\pm 0.004}$ | $0.973_{\pm 0.011}$ | $0.145_{\pm 0.031}$ | $1.027_{\pm 0.039}$ | | | |
| **PI-NN** | $\mathbf{0.992_{\pm 0.001}}$ | $\mathbf{0.982_{\pm 0.002}}$ | $\mathbf{0.120_{\pm 0.004}}$ | $1.046_{\pm 0.037}$ | | | |
| **GPR** | $0.993_{\pm 0.001}$ | $0.987_{\pm 0.002}$ | $0.102_{\pm 0.007}$ | $0.072_{\pm 0.003}$ | $94.33_{\pm 2.54}$ | $0.39_{\pm 0.01}$ | $1.44_{\pm 0.03}$ |
| **PI-GPR** | $\mathbf{0.995_{\pm 0.001}}$ | $\mathbf{0.990_{\pm 0.002}}$ | $\mathbf{0.091_{\pm 0.006}}$ | $\mathbf{0.064_{\pm 0.003}}$ | $93.85_{\pm 2.07}$ | $\mathbf{0.32_{\pm 0.01}}$ | $\mathbf{1.47_{\pm 0.03}}$ |
| **BNN -VI** | $0.984_{\pm 0.006}$ | $0.962_{\pm 0.014}$ | $0.172_{\pm 0.030}$ | $0.130_{\pm 0.017}$ | $95.46_{\pm 2.49}$ | $0.69_{\pm 0.04}$ | $1.09_{\pm 0.03}$ |
| **PI-BNN-VI** | $\mathbf{0.993_{\pm 0.001}}$ | $\mathbf{0.984_{\pm 0.003}}$ | $\mathbf{0.113_{\pm 0.008}}$ | $\mathbf{0.087_{\pm 0.006}}$ | $\mathbf{95.95_{\pm 1.01}}$ | $\mathbf{0.49_{\pm 0.02}}$ | $\mathbf{1.23_{\pm 0.03}}$ |
| **BNN-MCMC** | $\mathbf{0.996_{\pm 0.001}}$ | $\mathbf{0.991_{\pm 0.001}}$ | $\mathbf{0.085_{\pm 0.008}}$ | $\mathbf{0.060_{\pm 0.004}}$ | $92.71_{\pm 3.61}$ | $\mathbf{0.30_{\pm 0.03}}$ | $\mathbf{1.57_{\pm 0.07}}$ |
| **PI - BNN - MCMC** | $0.995_{\pm 0.001}$ | $0.990_{\pm 0.001}$ | $0.089_{\pm 0.002}$ | $0.064_{\pm 0.002}$ | $93.19_{\pm 1.21}$ | $0.32_{\pm 0.01}$ | $1.49_{\pm 0.03}$ |

Table 5 presents the results for the Nickel-based superalloys dataset, and Table 6 shows the results for the Titanium alloys dataset. The incorporation of physics knowledge led to significant improvements in the predictions for all models for these two datasets. The best-performing approach is PI-BB-MCMC for both datasets. As expected, PIML imparts greater benefits to tasks with smaller datasets, and we can see higher gains for these two datasets, compared to the physics-informed models with the larger SS316 alloys dataset.

*Table 5. Experimental results for the Nickel-based superalloys dataset with physics-informed features, including predictive accuracy metrics (PCC, $R^2$, RMSE, MAE) and uncertainty quantification metrics (coverage, interval width, composite metrics) for 8 compares methods: NN (Neural Network), QR (Quantile Regression), NGBoost (Natural Gradient Boosting), GPR (Gaussian Process Regression), Deep Ensemble, MC Dropout, BNN – VI (Variational Inference), and BNN – MCMC (Markov Chain Monte Carlo).*

|  | PCC ↑ | $R^2$ ↑ | RMSE ↓ | MAE ↓ | Coverage ↑ | Interval width ↓ | Composite metric ↑ |
|---|---|---|---|---|---|---|---|
| NN | 0.806$_{\pm 0.113}$ | **0.620**$_{\pm 0.190}$ | **0.241**$_{\pm 0.030}$ | **0.427**$_{\pm 0.060}$ |  |  |  |
| PI-NN | **0.905**$_{\pm 0.042}$ | 0.130$_{\pm 0.470}$ | 0.356$_{\pm 0.139}$ | 0.556$_{\pm 0.081}$ |  |  |  |
| GPR | **0.907**$_{\pm 0.019}$ | 0.801$_{\pm 0.054}$ | 0.175$_{\pm 0.026}$ | 0.125$_{\pm 0.011}$ | 92.77$_{\pm 5.56}$ | 0.594$_{\pm 0.019}$ | 1.12$_{\pm 0.048}$ |
| PI-GPR | 0.905$_{\pm 0.011}$ | **0.933**$_{\pm 0.021}$ | **0.102**$_{\pm 0.006}$ | **0.074**$_{\pm 0.007}$ | **94.73**$_{\pm 1.63}$ | **0.46**$_{\pm 0.05}$ | **1.40**$_{\pm 0.09}$ |
| BNN -VI | 0.909$_{\pm 0.032}$ | 0.791$_{\pm 0.069}$ | 0.185$_{\pm 0.030}$ | 0.147$_{\pm 0.025}$ | 87.52$_{\pm 5.80}$ | 0.534$_{\pm 0.029}$ | 1.15$_{\pm 0.044}$ |
| PI-BNN-VI | **0.972**$_{\pm 0.009}$ | **0.936**$_{\pm 0.015}$ | **0.101**$_{\pm 0.011}$ | **0.071**$_{\pm 0.007}$ | **97.38**$_{\pm 2.43}$ | **0.52**$_{\pm 0.02}$ | **1.24**$_{\pm 0.03}$ |
| BNN-MCMC | 0.914$_{\pm 0.027}$ | 0.824$_{\pm 0.050}$ | 0.167$_{\pm 0.023}$ | 0.116$_{\pm 0.019}$ | 92.84$_{\pm 4.71}$ | 0.607$_{\pm 0.061}$ | 1.12$_{\pm 0.021}$ |
| PI - BNN - MCMC | **0.992**$_{\pm 0.004}$ | **0.983**$_{\pm 0.009}$ | **0.049**$_{\pm 0.007}$ | **0.036**$_{\pm 0.003}$ | 93.46$_{\pm 5.80}$ | **0.20**$_{\pm 0.03}$ | **2.05**$_{\pm 0.05}$ |

*Table 6. Experimental results for the Titanium alloys dataset with physics-informed features, including predictive accuracy metrics (PCC, $R^2$, RMSE, MAE) and uncertainty quantification metrics (coverage, interval width, composite metrics) for 8 compares methods: NN (Neural Network), QR (Quantile Regression), NGBoost (Natural Gradient Boosting), GPR (Gaussian Process Regression), Deep Ensemble, MC Dropout, BNN – VI (Variational Inference), and BNN – MCMC (Markov Chain Monte Carlo).*

|  | PCC ↑ | R2 ↑ | RMSE ↓ | MAE ↓ | Coverage ↑ | Interval width ↓ | Composite Metric ↑ |
|---|---|---|---|---|---|---|---|
| NN | 0.794$_{\pm 0.227}$ | **0.381**$_{\pm 0.777}$ | **0.800**$_{\pm 0.481}$ | 1.326$_{\pm 0.150}$ |  |  |  |
| PI-NN | **0.851**$_{\pm 0.095}$ | -0.119$_{\pm}$ | 1.079$_{\pm 1.723}$ | 1.606 |  |  |  |
| GPR | 0.921$_{\pm 0.034}$ | 0.839$_{\pm 0.066}$ | 0.471$_{\pm 0.081}$ | 0.309$_{\pm 0.044}$ | 93.24$_{\pm 2.22}$ | 1.99$_{\pm 0.20}$ | 0.87$_{\pm 0.017}$ |
| PI-GPR | **0.941**$_{\pm 0.027}$ | **0.877**$_{\pm 0.054}$ | **0.433**$_{\pm 0.118}$ | **0.298**$_{\pm 0.076}$ | **93.46**$_{\pm 6.85}$ | **1.82**$_{\pm 0.31}$ | **0.88**$_{\pm 0.03}$ |
| BNN -VI | 0.919$_{\pm 0.039}$ | 0.811$_{\pm 0.116}$ | **0.474**$_{\pm 0.096}$ | 0.320$_{\pm 0.040}$ | 75.10$_{\pm 9.36}$ | 0.90$_{\pm 0.06}$ | 0.86$_{\pm 0.091}$ |
| PI-BNN-VI | **0.934**$_{\pm 0.015}$ | **0.836**$_{\pm 0.051}$ | 0.503$_{\pm 0.063}$ | 0.353$_{\pm 0.027}$ | 73.16$_{\pm 5.37}$ | 0.96$_{\pm 0.05}$ | 0.84$_{\pm 0.05}$ |
| BNN-MCMC | 0.922$_{\pm 0.029}$ | 0.843$_{\pm 0.061}$ | **0.449**$_{\pm 0.050}$ | 0.336$_{\pm 0.023}$ | **94.87**$_{\pm 4.21}$ | 1.77$_{\pm 0.14}$ | 0.86$_{\pm 0.026}$ |
| PI - BNN - MCMC | **0.937**$_{\pm 0.012}$ | **0.865**$_{\pm 0.031}$ | 0.465$_{\pm 0.082}$ | **0.288**$_{\pm 0.029}$ | 86.90$_{\pm 6.32}$ | **1.09**$_{\pm 0.32}$ | **0.98**$_{\pm 0.15}$ |

*4.4.3 Active Learning*

For the AL case study, we used a Batch Mode AL with a batch size $B$ of 10 for the SS316 alloy dataset and a batch size $B$ of 8 for the smaller Nickel-based superalloy and Titanium alloy datasets. Figure 4 presents the plots for PCC and $R^2$ scores of the GPR, BNN-VI, and BNN-MCMC models. For the SS316 dataset with PI features, GPR achieves high PCC and $R^2$ scores with fewer data points. For Nickel and Titanium alloys, BNN MCMC achieves the best result, and it converges the fastest with a small number of data samples. Overall, BNN-MCMC performed the best in 2 of the 3 tests and GPR performed the best in 1 of the tests, whereas BNN-VI had the lowest performance overall.

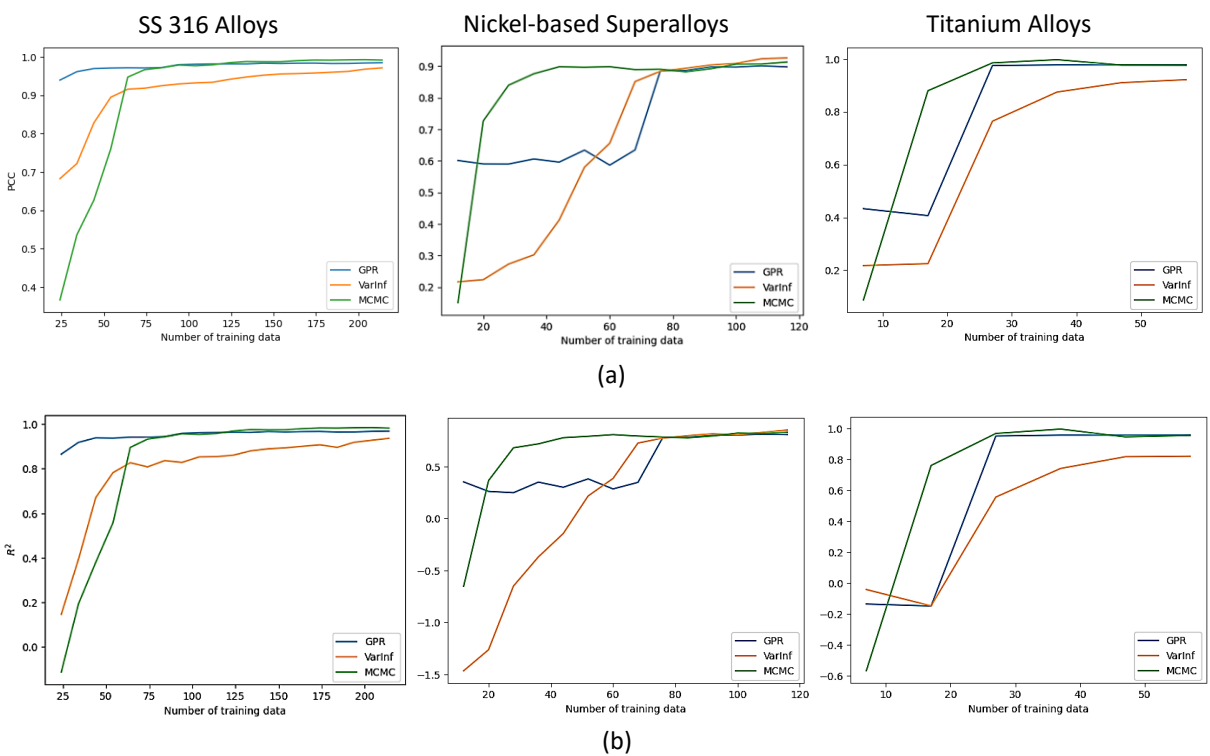

*Figure 4.* (a) PCC and (b) $R^2$ score BMAL plots with Gaussian Process Regression (GPR), BNN-Variational Inference, and BNN-Markov Chain Monte Carlo (MCMC) for SS316 alloys, Nickel-based superalloys, and Titanium alloys.

## 5. Discussion

Our experimental results in Tables 1 to 3 indicate that BNN models and GPR outperform the traditional ML models, and therefore they are more suitable for UQ in material property prediction. Overall, BNN-MCMC achieved the best performance on most metrics. The results in Tables 4 to 6 demonstrate that the PIML paradigm can improve the models' performance for the smaller Nickel-based superalloy and Titanium alloy datasets, whereas it obtained smaller improvement for the larger SS316 alloys dataset. For the application of UQ in AL, BNN-MCMC and GPR performed comparatively well.

The experimental results demonstrate the capability of BNNs to provide accurate and reliable UQ. Bayesian NNs exhibit more consistent uncertainty that aligns better with the observed deviations, reducing the likelihood of overconfidence or underconfidence. Additionally, BNNs allow for the separation of epistemic uncertainty and aleatoric uncertainty. This property enhances the data efficiency of BNNs, enabling effective learning from smaller datasets. Accordingly, the priors in BNNs can be regarded as soft constraints that act similar to the regularization techniques in traditional NNs. Likewise, physics-informed BNNs leverage physical knowledge to constrain and guide the learning process, and it can also be used to generate additional features when dealing with insufficient data. AL actively selects the most informative samples to ensure the models learn faster with fewer samples.

One major limitation of BNNs is their high computational cost, since both BNN-MCMC and BNN-VI require sampling from the posterior distribution over the network parameters. Furthermore, MCMC requires sufficient number of iterations to obtain accurate samples from the posterior distribution, which can take significantly longer than GPR and require increased computational resources. Also, hyperparameter tuning of BNNs can be challenging and requires a deeper understanding of the internal working of the models.

It is also worth noting that GPR primarily focuses on modeling structural uncertainty reflecting the inherent variability within the model structure, as the function space is modeled as a distribution over functions, and the predictions are made by considering the possible functions that are consistent with the observed data. On the other hand, BNNs are aimed at quantifying parametric uncertainty that arises from the variability in the model parameters, rather than uncertainty in the functional form of the model.

A limitation of this work is that the collected data from creep rupture tests do not comprise data from repeated experimental measurements that capture the variability in the measured creep life values. Consequently, the used datasets do not provide ground truth values to allow for truthful evaluation and comparison of the used methods for uncertainty quantification.

In future work, we will focus on the development of physics-informed loss functions and physics-informed layers in BNNs, and implementation of AL approaches based on hybrid query strategies, such as Query by Committee.

## 6. Conclusion

This work provides a study of uncertainty quantification in multivariable regression for material property prediction with Bayesian Neural Networks. In our proposed approach, we developed a PIML framework for predicting creep rupture life in metal alloys by introducing physics-informed feature engineering to augment the set of input features to the regression models and by designing a physics-informed loss function that introduces physics constraints into the learning algorithm. The hypothesis we test is that the synergistic combination of historical experimental data from creep tests and prior knowledge and constraints from the governing physics laws into a PIML framework will yield improved predictions and uncertainty quantifications of creep deformation, that satisfy physics constraints and are consistent with the underlying physics laws. Our findings demonstrate the potential of BNNs to advance the field of materials science and engineering by enabling more accurate and reliable predictions with quantified uncertainties. The experimental validation indicates that the most promising approach for material property prediction is BNN-MCMC, which achieved performance that is either competitive or exceeds the performance of GPR as the state-of-the-art method for UQ is multivariable regression. The case study on applying uncertainty estimates in an active learning scenario confirms that BNN is a promising approach for overcoming the challenges in modeling material properties related to sparse and noisy data.

**Acknowledgments**

This work was supported by the University of Idaho—Center for Advanced Energy Study (CAES) Seed Funding FY23 Grant. This work was also supported through the INL Laboratory Directed Research & Development (LDRD) Program under DOE Idaho Operations Office Contract DE-AC07-05ID14517 (project tracking number 23A1070-069FP). Accordingly, the publisher, by accepting the article for publication, acknowledges that the U.S. Government retains a nonexclusive, paid-up, irrevocable, worldwide license to publish or reproduce the published form of this manuscript or allow others to do so, for U.S. Government purposes.